\pdfoutput=1

\documentclass[11pt]{article}

\usepackage{emnlp2021}

\usepackage{times}
\usepackage{latexsym}
\usepackage{graphicx} 
\usepackage{booktabs} 
\usepackage{placeins} 
\usepackage{microtype} 
\usepackage{enumitem} 
\usepackage{amsmath, bm} 
\usepackage[skip=2pt]{subcaption} 
\usepackage{comment} 
\usepackage[labelfont=bf]{caption} 

\usepackage{color, colortbl}
\usepackage[T1]{fontenc}

\usepackage[utf8]{inputenc}

\usepackage{microtype}

\let\SUP\textsuperscript

\title{Namesakes: \\
Ambiguously Named Entities from Wikipedia and News}

\author{Oleg Vasilyev, Aysu Altun, Nidhi Vyas\thanks{* Currently at Galileo Technologies}, Vedant Dharnidharka, Erika Lam, John Bohannon \\
  Primer Technologies Inc. \\
  San Francisco, California \\
  \texttt{\{oleg,aysu.altun,erika.lam,vedant,john\}@primer.ai}\\}

\date{}

\begin{document}
\maketitle
\begin{abstract}
We present Namesakes, a dataset of ambiguously named entities obtained from English-language Wikipedia and  news articles. It consists of 58862 mentions of 4148 unique entities and their namesakes: 1000 mentions from news, 28843 from Wikipedia articles about the entity, and 29019 Wikipedia backlink mentions. Namesakes should be helpful in establishing challenging benchmarks for the task of named entity linking (NEL).
\end{abstract}

\section{Introduction}
Recent advances have made it possible to incorporate knowledge into distributed neural representations \cite{Min2020knowledge, nooralahzadeh-ovrelid-2018-sirius}.
A fundamental component of such systems is named entity linking (NEL) \cite{yang2016smart, sorokin-gurevych-2018-mixing, kolitsas2018endtoend, li2020efficient, sevgili2021neural}. Given a text, the task is to  correctly identify mentions of named entities by linking to the correct reference 
entities in a knowledge base, e.g. Wikipedia. As the world evolves, new entities and new information about existing entities must be tracked with a dynamic knowledge base.

If every entity had a unique name like a bar code, NEL would be easy. But we live in a world populated by "Michael Jordan", a name shared by a renowned computer scientist, a famous basketball player, a famous actor, and many others. There are more than 20 entities with the surface form "Michael Jackson" in Wikipedia\footnote{en.wikipedia.org/wiki/Michael\_Jackson\_(disambiguation)}, and the Wikipedia Disambiguation pages for some names include hundreds of unique entities.\footnote{en.wikipedia.org/wiki/Aliabad} 

In some definitions of NEL the task includes named entity recognition (NER), the initial tagging of named entity mentions in the target text \cite{yang2016smart, sorokin-gurevych-2018-mixing, kolitsas2018endtoend, li2020efficient, sevgili2021neural}. Here we focus on the more narrow sense of NEL which assumes the initial tagging of named entities is done \cite{rao2012entity, Wu2020ScalableZE, Logeswaran2019ZeroShotEL}. The mentions of entities in a text corpus are generally provided as spans that capture various surface forms. For example for the entity \textit{Michael Jordan}, some mentions include \textit{M. Jordan, Jordan, Michael Jordan}. 

Most existing NEL-related datasets do not focus on highly ambiguous names, e.g. WikiDisamb30 \cite{Ferragina2012FastAA}, ACE and MSNBC \cite{ratinov-etal-2011-local}, WNED-CWEB and WNED-WIKI \cite {guo2018} CoNLL-YAGO \cite{hoffart-etal-2011-robust}, and TAC KBP Entity Discovery and Linking dataset \cite{Ji2017OverviewOT}. The recently introduced Ambiguous Entity Retrieval (AmbER) dataset by \citet{chen2021evaluating} is an exception, including subsets of identically named entities for the purpose of fact checking, slot filling, and question-answering tasks. AmBer is limited to Wikipedia text and was automatically generated. 

The unique contribution of the \textbf{Namesakes} \cite{namesakes2021} dataset\footnote{figshare.com/articles/dataset/Namesakes/17009105/1} is its diversity—it includes news mentions—and its high quality, ensured by manual data-labeling. The importance of manual labeling will become clear in the data description that follows. 
In this paper we describe in detail the data selection, filtering, and composition of Namesakes. We also define and present the ambiguity of the mentions of entities in Namesakes.
The aim of Namesakes is to help researchers distinguish the performance of NEL systems with highly challenging, realistic data.

\section{Dataset composition}
\subsection{Data motivation}
The primary motivation for this work is to create a more challenging NEL dataset where:
\begin{enumerate}[topsep=0pt,itemsep=-1ex,partopsep=1ex,parsep=1ex]
    \item The names of most entities must be ambiguous, i.e. with Wikipedia disambiguation pages linking to multiple Wikipedia articles.
    \item Most entities must belong to one of three categories - person, location, organization - preferably in balanced proportions.
    \item Entity contexts are captured by three distinct components: 
    \begin{enumerate}
        \item \textit{Entities}: Wikipedia articles describing named entities
        \item \textit{Backlinks}: Wikipedia articles that reference members of \textit{Entities}
        \item \textit{News}: News articles containing mentions from \textit{Entities}. The News mentions may be true references to members of \textit{Entities} or have name collisions with them.
    \end{enumerate}
    \item All three components should have reasonably clean text chunks (at least in the neighborhood of the entities of interest), filtered for obscure reference sections, business reports, TV listings, etc.
\end{enumerate}
The initial entities names we selected are listed in the Appendix \ref{sec:Ambiguous_Names}. Only names with a Wikipedia disambiguation page are included, resulting in an initial count of 7626 Wikipedia entities.

\subsection{Entities: selection for labeling}\label{sec:Selection_for_labeling_entities}
We filtered the preliminary 7626 Wikipedia entities by requiring that each text satisfies several reasonable conditions:
\begin{enumerate}[topsep=0pt,itemsep=-1ex,partopsep=1ex,parsep=1ex]
    \item The text must contain at least 5 "good sentences", after sentence tokenization with NLTK. The "good sentence" empirical conditions are listed in Appendix \ref{sec:Good_sentences}. 
    \item The first named entity identified in the text by an NER model must be a person PER, organization ORG or location LOC, not miscellaneous MISC. We performed NER using the  "bert-large-cased-finetuned-conll03-english" model from the Hugging Face model hub\footnote{https://huggingface.co/dbmdz/bert-large-cased-finetuned-conll03-english}.
    \item The text must contain at least 3 mentions that are similar to the Wikipedia entity's name. These entities should be picked up only from the "good sentences". The similarity here means that an entity contains at least one word from the entity name with length >=3 characters. The motivation for this is to exclude abnormally short Wikipedia entities, or at least have an entity containing entities that can be confused with the "main" entity.  
\end{enumerate}

With the highly ambiguous entities we started with, and after the above filtering, the resulting Wikipedia pages had a high frequency of intentionally confusing entities. Most of the ambiguity derives from having similar or identical names shared among multiple Wikipedia entities. Additional confusion comes from a Wikipedia entity text frequently having not only the "main" entity (the main focus of the article), but additional entities with the same or similar names described in the article. These additional entities are often relatives of a "main" person entity, or locations and organizations of the same or similar name. 


The goal of our labeling was to identify in each entity the "main" entity and the "other" entities. In our final selection for labeling, we selected top entity names by the number of Wikipedia pages in which such names occurred. Specifically, the top 100 names of each type were selected: person, organization, location. Our resulting dataset for labeling contains 4148 Wikipedia entities. For each entity text, all the entities with the names similar to the name of the entity were tagged as "\textit{categorize}", requesting annotators to replace the tag by either "\textit{Same}" or "\textit{Other}". The tag "Same" means that the entity is actually the entity, i.e. the entity to which the entity text is devoted to. The tag "Other" means that, despite the same or similar name, the tagged entity is not the entity.

\subsection{Entities data}
The labeled Entities data is the output of our manual labeling process, undertaken by an annotation team at Odetta\footnote{https://odetta.ai} using the annotation software tool Datasaur\footnote{https://datasaur.ai}.
The team consisted of 6 annotators who were experienced with NLP projects and passed a trial training task for this project. Only the most reliable tagged entity mentions were kept:
\begin{enumerate}[topsep=0pt,itemsep=-1ex,partopsep=1ex,parsep=1ex]
    \item The mentions to which all six annotators assigned the "Same" tag.
    \item The mentions on which only one annotator disagreed with the rest; such mentions were confirmed via reconciliation. 
\end{enumerate}
The final result was 21426 "Same" mentions and 7417 "Other" mentions for the 4148 Wikipedia entities.

\subsection{News data}\label{sec:news}
The News component of our dataset was created from querying Primer's proprietary news corpus by the ambiguous entity names from the Entities component. The "ambiguous" entities are the entities with the names (aliases, or last names of people) that could be mixed up with at least three other entities. Once the news articles were obtained, they were filtered (see Appendix \ref{sec:News_filter}) by excluding articles not satisfying the requirements of text quality and  manageable labeling:
\begin{enumerate}[topsep=0pt,itemsep=-1ex,partopsep=1ex,parsep=1ex]
    \item The final number of news texts is 1000.
    \item Each text is between 500 and 3000 characters.
    \item Each text has a named entity mention found by the name query; this mention has to be labeled. The list of suggested labels must contain at least 3 but not more than 10 Wikipedia entities with which the mention can be confused, including the entity (if it exists) to which it belongs. Fewer than 3 Wikipedia entities fails to provide enough ambiguity, while more than 10 would be too time-consuming to label.
\end{enumerate}
The goal of the labeling was to assign to each mention its correct Wikipedia entity (if existing in the Entities dataset), from the list of 3-10 provided Wikipedia entities. The labeling was done using the annotation tool Datasaur, by 3 Odetta annotators, with consequent reconciliation of all the mentions that caused a disagreement. Similar to the labeling of the Entities, the annotators were experienced in NLP projects and went through a trial task.

The resulting News data consists of 1000 texts, each text with one annotated mention. Of these mentions, 276 do exist in the Entities data, and 724 do not exist in the Entities data (but can be easily confused with many entities from there).

\subsection{Backlinks data}\label{sec:Backlinks}
The \textit{Backlinks} dataset was manufactured from a Wikipedia dump from July 1, 2021 by collecting (and then thoroughly filtering for quality) the entities that link to the \textit{Entities} dataset. For example, if the person described in the "John Muir" Wikipedia article were a member of the Entities dataset, then the "National Park" Wikipedia page would be one of our candidate backlinks, because its text includes a hyperlink with anchor text "John Muir" that links to the "John Muir" page. 

Many links happen to occur in reference-like sections of Wikipedia rather than in a normal text; this requires careful filtering. Our filtering and cleaning included (for details see Appendix \ref{sec:Backlinks_filter}):
\begin{enumerate}[topsep=0pt,itemsep=-1ex,partopsep=1ex,parsep=1ex]
    \item Removing Wikipedia pages with titles that start with certain words, such as "List", "MediaWiki" etc.
    \item Removing bottom part of the page text, starting from certain sections, named like "Notes, "References" etc.
    \item Only mentions from "good" parts of the Wikipedia page are kept. If a page loses all its mentions, the page is removed.
    \item Any page text is cut after 1000 characters down from the last (occurred lowest in the text) mention. 
\end{enumerate}
The resulting Backlinks dataset contains 26903 text chunks from Wikipedia pages, and 29019 linked mentions in the pages.

\subsection{Resulting dataset}\label{sec:Resulting_Dataset}
\begin{figure}[th]
\includegraphics[width=0.48\textwidth]{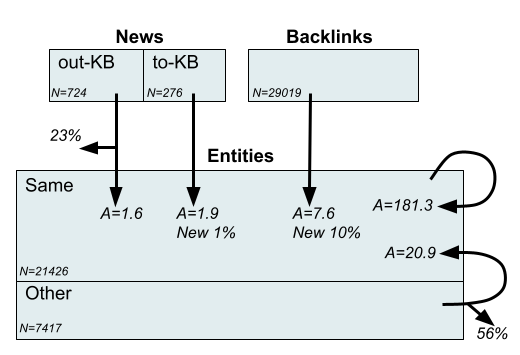}
\caption{Structure of Namesakes. An attempt to link a mention to KB (Entities with "Same" mentions) has a potential confusion - Ambiguity $A$ (discussed in Section \ref{sec:DatasetFeatures} and defined in Appendix \ref{sec:Overlaps}). $N$ is the number of mentions. Some mentions that refer to KB entities have mentions not existing in the KB, the percent of such mentions is indicated as "New". Some mentions that should not belong to KB ("out-KB" News and "Other" Entities) nevertheless have namesakes of "Same" mentions - as shown by bifurcated arrows. This figure shows only one possible scenario of possible linking, see Section \ref{sec:Discussion}.}
\label{fig:Namesakes}
\end{figure}

The resulting dataset Namesakes consists of three closely related datasets: Entities, News, and Backlinks. The structure of the dataset is shown in Figure \ref{fig:Namesakes}, the details of the figure will be explained in the following subsections and in Section \ref{sec:DatasetFeatures}.
The Entities and Backlinks consist of Wikipedia text chunks. The News consists of random news chunks. The Entities and News are human-labeled, resolving the mentions of the entities. The Backlinks are not labeled, but have mentions already linked by Wikipedia. In this section we summarize the structure of the data.

\subsubsection{Entities}
The Entities dataset consists of 4148 Wikipedia text chunks containing human-tagged mentions of entities. Each mention is tagged either as "Same" (meaning that the mention is of this Wikipedia page entity), or "Other" (meaning that the mention is of some other entity, just having the same or similar name). The Entities dataset is a jsonl list, each item is a dictionary with the following keys and values:
\begin{enumerate}[topsep=0pt,itemsep=-1ex,partopsep=1ex,parsep=1ex]
    \item Key "pagename": page name of the Wikipedia page.
    \item Key "pageid": page id of the Wikipedia page.
    \item Key "title": title of the Wikipedia page.
    \item Key "url": URL of the Wikipedia page.
    \item Key "entities": list of the mentions in the page text, each entity is represented by a dictionary with the keys: 
    \begin{enumerate}[topsep=0pt,itemsep=-1ex,partopsep=1ex,parsep=1ex]
        \item Key "text": the mention as a string from the page text.
        \item Key "start": start character position of the entity in the text.
        \item Key "end": end (one-past-last) character position of the entity in the text.
        \item Key "tag": the annotation tag ("Same" or "Other") given to the mention.
    \end{enumerate}
    \item Key "text": The text chunk.
\end{enumerate}
The texts contain 21426 mentions tagged "Same", and 7417 mentions tagged "Other".

\subsubsection{News}
The News dataset consists of 1000 news text chunks, each with a single annotated entity mention. The annotation either points to the corresponding entity from the Entities dataset (if the mention is of that entity), or indicates that the mention entity does not belong to the Entities dataset. The News dataset is a jsonl list, each item is a dictionary with the following keys and values:
\begin{enumerate}[topsep=0pt,itemsep=-1ex,partopsep=1ex,parsep=1ex]
    \item Key "id\_text": Id of the document (0,1,2,3,...).
    \item Key "entity": a dictionary describing the annotated entity mention in the text:
    \begin{enumerate}[topsep=0pt,itemsep=-1ex,partopsep=1ex,parsep=1ex]
        \item Key "text": the mention as a string found by an NER model in the text.
        \item Key "start": start character position of the mention in the text.
        \item Key "end": end (one-past-last) character position of the mention in the text.
        \item Key "tag": This key exists only if the mentioned entity is annotated as belonging to the Entities dataset - if so, the value is a dictionary identifying the Wikipedia page assigned by annotators to the mentioned entity:
        \begin{enumerate}[topsep=0pt,itemsep=-1ex,partopsep=1ex,parsep=1ex]
            \item Key "pageid": Wikipedia page id.
            \item Key "pagetitle": page title.
            \item Key "url": page URL.
        \end{enumerate}
    \end{enumerate}
    \item Key "urls": List of URLs of wikipedia entities suggested to labelers for identification of the entity mentioned in the text.
    \item Key "text": The text chunk.
\end{enumerate}
Of the 1000 mentions, 276 do exist in the Entities dataset, and the rest do not (but have the names that could be easily confused with one or more entities from there).

\subsubsection{Backlinks}
The Backlinks dataset consists of two parts: dictionary Entity-to-Backlinks and Backlinks documents. The dictionary points to backlinks for each entity of the Entity dataset (if any backlinks exist for the entity). The Backlinks documents are the backlinks Wikipedia text chunks with identified mentions of the entities from the Entities dataset. 

Each mention is identified by surrounded double square brackets, e.g. "Muir built a small cabin along [[Yosemite Creek]].". However, if the mention differs from the exact entity name, the double square brackets wrap both the exact name on the left and, separated by '|', the mention string on the right, for example: "Muir also spent time with photographer [[Carleton E. Watkins | Carleton Watkins]] and studied his photographs of Yosemite.". 

The Entity-to-Backlinks is a jsonl with 1527 items, each item is a tuple:
\begin{enumerate}[topsep=0pt,itemsep=-1ex,partopsep=1ex,parsep=1ex]
    \item Entity name.
    \item Entity Wikipedia page id.
    \item Backlinks ids: a list of pageids of backlink documents.
\end{enumerate}

The Backlinks documents is a jsonl with 26903 items, each item is a dictionary:
\begin{enumerate}[topsep=0pt,itemsep=-1ex,partopsep=1ex,parsep=1ex]
    \item Key "pageid": Id of the Wikipedia page.
    \item Key "title": Title of the Wikipedia page.
    \item Key "content": Text chunk from the Wikipedia page, with all mentions in the double brackets; the text is cut 1000 characters after the last mention, the cut is denoted as "...[CUT]".
    \item Key "mentions": List of the mentions from the text, for convenience. Each mention is a tuple:
    \begin{enumerate}[topsep=0pt,itemsep=-1ex,partopsep=1ex,parsep=1ex]
        \item Entity name.
        \item Entity Wikipedia page id.
        \item Sorted list of all character indexes at which the mention occurrences start in the text.
    \end{enumerate}
\end{enumerate}

\section{Dataset features}\label{sec:DatasetFeatures}
\subsection{Entities dataset}\label{sec:DatasetFeatures_Entities}
\begin{figure}[th]
\includegraphics[width=0.45\textwidth]{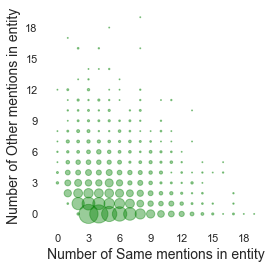}
\caption{In Entities: Distribution of entities by the number of the mentions in them. Area of (X,Y) marker is proportional to the number of entities having X "Same" and Y "Other" mentions.}
\label{fig:Entity_scatter}
\end{figure}

\begin{figure}[th]
\includegraphics[width=0.48\textwidth]{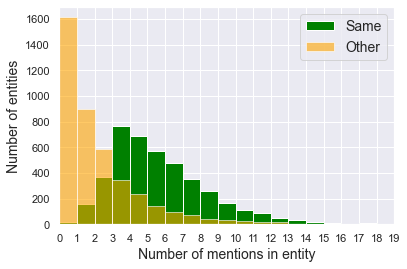}
\caption{In Entities: Histogram of entities by the number of the mentions in entity. Two entities (having 23 mentions and 27 mentions) are cut from the histogram. Average number of "Same" mentions per entity is 5.2.}
\label{fig:Entity_mentions}
\end{figure}

The Entities dataset by itself is simple: its 4148 Wikipedia text chunks contain 28843 annotated mentions, of which 21426 are "Same" and 7417 are "Other". 
The 21426 "Same" mentions are occurrences of 2909 unique mentions as strings. The 7417 "Other" mentions are occurrences of 3754 unique mentions as strings. The distribution of the entities by the number of the mentions they have is shown in Figures \ref{fig:Entity_scatter} and \ref{fig:Entity_mentions}.

The 4148 entities of Entities dataset have high overlap of their names. For example, if we define an "overlap" as two names having at least one common word of at least 4 characters, then there are 4132 overlapping entities out of 4148. The most overlapping entity is \textit{"John William Smith (legal writer)"}, - overlaps with 1241 other entities. In our definition of the overlap we excluded all the bracketed categorizations, like \textit{"(legal writer)"}, because such words are generic and not really parts of the name. We can characterize the confusing potential of a dataset by a \textbf{names overlap}, which we define as the number of entities with which an average entity overlaps by at least one non-generic word of length >= 4 characters. For Entities dataset we have \textit{names overlap = 379.5}.

\begin{figure}[th]
\includegraphics[width=0.48\textwidth]{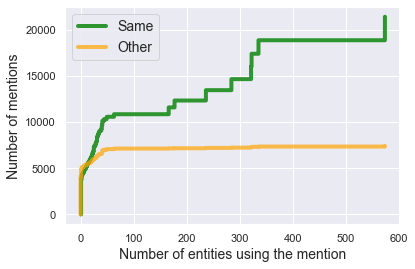}
\caption{In Entities: cumulative distribution of mentions by the number of entities using the mention as "Same" (see Appendix \ref{sec:Overlaps_Entities}). On the right the total number of mentions reaches 21426 for "Same, and 7417 for "Other". The average ambiguities is 181.3 for "Same" and 20.9 for "Other".}
\label{fig:Entity_ambiguity}
\end{figure}

The real problem for NEL happens when the same mention was used legitimately for more than one entity. We can characterize potential confusion of NEL directly in terms of the mentions. How many entities an average mention may refer to? More specifically, what is the average number of Wikipedia entities that had at least one "Same" mention coinciding with the considered mention? This characteristic, \textbf{mention-entities ambiguity}, equals 181.3 for the mentions "Same", and it equals 20.9 for the mentions "Other" (see Appendix \ref{sec:Overlaps_Entities}).
This makes a strong potential for confusing NEL. The distribution of individual ambiguities is shown in Figure \ref{fig:Entity_ambiguity}.

\subsection{News - Entities}
The News dataset contains 1000 text chunks, each with one mention annotated either as belonging to the Entities dataset (276 cases), or not (724 cases). We will call these mentions "to-KB" and "out-KB" mentions correspondingly. The 276 mentions recognized as belonging to the Entities are actually occurrences of 24 unique mentions. The 724 mentions recognized as not belonging to the Entities are occurrences of 54 unique mentions.

A strong possibility of confusion for NEL comes from the fact that some unique mentions from News completely coincide with the "Same" mentions from more than one entity from the Entities.
Even stronger confusion comes from the News mentions that do not belong to the Entities, but nevertheless exactly coincide with the "Same" mentions from the Entities. Of 54 unique mentions recognized as not belonging to Entities, 37 coincide with at least one "Same" mention from the Entities. This makes 44\% of the "Other" mentions occurrences; the remaining 56\%, while having some names overlap, do not exactly coincide with any "Same" mentions or with exact entity names, - as depicted in Figure \ref{fig:Namesakes}.
 
\begin{figure}[th]
\includegraphics[width=0.48\textwidth]{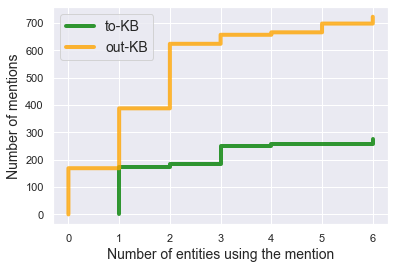}
\caption{News to Entities: cumulative distribution of mentions by the number of entities using the mention as "Same" (see Appendix \ref{sec:Overlaps_News}). On the right the total number of mentions reaches 276 for "to-KB", and 724 for "out-KB". The corresponding average ambiguities are 1.9 and 1.6.}
\label{fig:News_ambiguity}
\end{figure}

The \textit{mention-entities ambiguity} is 1.9 for the 276 "to-KB" News mentions, and it is 1.6 for the 724 "out-KB" News mentions (see Appendix \ref{sec:Overlaps_News}). The distribution of individual ambiguities is shown in Figure \ref{fig:News_ambiguity}.

\subsection{Backlinks - Entities}
The relation Backlinks-Entities is simpler than the relation News-Entities, because all the mentions in Backlinks do belong to the Entities (and there was no need to annotate). All 29019 mention occurrences in Backlinks refer to 1527 entities of the Entities dataset. Since the mention "surface forms" do not always coincide with the entity names, there are 2399 (rather than 1527) unique mentions in the Backlinks dataset. We depicted in Figure ~\ref{fig:Namesakes} that 10\% of the mention occurrences do not coincide with any "Same" mentions or entity names of Entities. 

\begin{figure}[th]
\includegraphics[width=0.48\textwidth]{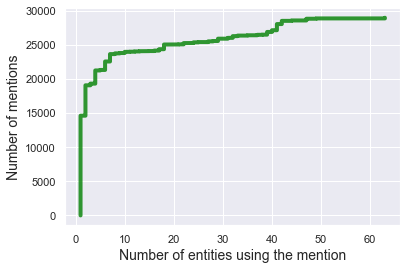}
\caption{Backlinks to Entities: cumulative distribution of Backlinks mentions by the number of entities using the mention as "Same" in Entities. On the right of the plot the total number of mentions reaches 28991 (the remaining 28 mentions are cut from the plot, their number of entities is in the range 100 - 600). The average mention-entities ambiguity is 7.6. }
\label{fig:Backlinks_ambiguity}
\end{figure}

The \textit{mention-entities ambiguity} equals 7.6 for Backlinks mentions (the ambiguity is with respect to the Entities, see Appendix \ref{sec:Overlaps_News}). The distribution of individual ambiguities is shown in Figure \ref{fig:Backlinks_ambiguity}.

\section{Discussion}
\label{sec:Discussion}
The Namesakes dataset is helpful to us for evaluating NEL models, and we hope it will be useful as well for the community. High ambiguity makes it easier to distinguish between otherwise similar performance. There are several scenarios for using Namesakes for NEL evaluation, depending on the role played by a knowledge base (KB), and what role is played by external mentions (EM) that must be linked to the KB.

NEL for an EM mention may either link to a KB entity, or not link to the KB at all. The first outcome is incorrect in two cases: the mention refers to an entity which is not in KB, or the linking was done to an incorrect KB entity. The second outcome is incorrect if the mention refers to an entity existing in the KB (we will call such a mention a \textit{to-KB mention}, and we will call any other mention an \textit{out-KB mention}). 

Problems for an NEL model may come not only from the ambiguity of the EM mentions with respect to the KB, but also from the difference in style of the EM texts versus KB texts. In Namesakes, Entities texts are written in clear Wikipedia style; News texts are more varied and may have much less context relevant to the mention; and Backlinks texts are in between: still in the Wikipedia style, but with context not centered around the mention.  

A few simple scenarios of splitting Namesakes on EM and KB: 
\begin{enumerate}[topsep=0pt,itemsep=-1ex,partopsep=1ex,parsep=1ex]
    \item News to Entities
    \item Backlinks to Entities
    \item News to Entities+Backlinks
    \item Entities to Entities.
    \item All to All
\end{enumerate}
The News-to-Entities scenario is relevant for evaluating NEL for a situation where the KB was created from knowledge-friendly texts (in our case Entities is from Wikipedia), and EM come from accidental encounters with named entities in varied sources (in our case News texts). As in most scenarios, the evaluation must include three numbers:
\begin{enumerate}[topsep=0pt,itemsep=-1ex,partopsep=1ex,parsep=1ex]
    \item For to-KB mentions:
    \begin{enumerate}[topsep=0pt,itemsep=-1ex,partopsep=1ex,parsep=1ex]
        \item Percent of mentions linked to a wrong KB entity
        \item Percent of mentions not linked to KB
    \end{enumerate}
    \item For out-KB mentions: percent of mentions linked to KB
\end{enumerate}
Arguably, it is most important to have the third listed percent low, almost as important to have the first one low, and less important but desirable to have the second percent low.  

The Backlinks-to-Entities scenario still provides some difference between the EM texts style and the KB texts style, because a backlink text mentions the entity of interest only incidentally. The Backlinks dataset has only to-KB mentions, so for simulating out-KB mentions it is necessary to exclude some entities from the Entities.

The News-to-Entities+Backlinks scenario assumes that the KB is created both from the Entities and the Backlinks, and thus should be better for linking from varied texts like News. An NEL system would be expected to perform in this scenario better than in the News-to-Entities.   

The Entities-to-Entities addresses a situation where EM texts are similar to the KB texts, and provides very high ambiguity of the entities. However, it is necessary to carefully split the Entities into the mentions and entities for EM and the mentions and entities for KB.

The All-to-All scenario assumes 'All = Entities + Backlinks + News', i.e. the whole Namesakes dataset must be split on EM and KB. This allows insights on how different kinds of mentions can be useful for creating KB entities, and how easy it is to link different kinds of EM mentions.

\section{Conclusion}
Advances in named entity linking (NEL) require a sensitive test of performance to distinguish good from great performance. To that end, we created Namesakes \cite{namesakes2021}, a dataset\footnote{figshare.com/articles/dataset/Namesakes/17009105/1} of entities with highly ambiguous names. In this paper we outline our motivation and method for data selection, filtering, and composition. We also describe in detail and define the metrics of ambiguity for these entities.  

\section*{Acknowledgments}
We thank Spencer Braun for review of the paper and valuable feedback; we thank Odetta\footnote{https://odetta.ai} annotators and reconciliators, especially Nusaiba Khubaib, for high dedication to managing the labeling and verifying the data.

\bibliography{anthology,custom}
\bibliographystyle{acl_natbib}

\appendix

\section{Ambiguous Names}\label{sec:Ambiguous_Names}
Table \ref{tab:names_people} contains list of common person names that we used. The names were taken from the top of the most common names lists
\footnote{\url{https://www.ssa.gov/oact/babynames/decades/century.html}}
\footnote{\url{List_of_most_common_surnames_in_North_America##United_States_(American)}}

\begin{table}[th]
\caption{Common person names: 10 most common male first names (column 1), female first names (column 2) and last names (column 3).}
\centering
\begin{tabular}{@{}lcccc@{}}
\toprule
\textbf{male} & \textbf{female} & \textbf{surname}\\
\hline
James & Mary & Smith\\
Robert & Patricia & Johnson\\
John & Jennifer & Williams\\
Michael & Linda & Brown\\
William & Elizabeth & Jones\\
David & Barbara & Miller\\
Richard & Susan & Davis\\
Joseph & Jessica & Garcia\\
Thomas & Sarah & Rodriguez\\
Charles & Karen & Wilson\\
\bottomrule
\end{tabular}
\label{tab:names_people}
\end{table}

Combining first and last names from the table gives 200 person names.

List of 35 common location names that we used (from the most common US locations\footnote{\url{List_of_the_most_common_U.S._place_names}}): 
'Washington', 'Franklin', 'Arlington', 'Centerville', 'Lebanon', 'Clinton', 'Springfield', 'Georgetown', 'Fairview', 'Greenville', 'Bristol', 'Chester', 'Dayton', 'Dover', 'Madison', 'Salem', 'Oakland', 'Milton', 'Newport', 'Riverside', 'Ashland', 'Bloomington', 'Manchester', 'Oxford', 'Winchester', 'Burlington', 'Jackson', 'Milford', 'Clayton', 'Mount Vernon', 'Auburn', 'Kingston', 'Lexington', 'Cleveland', 'Hudson'.

Initial list of organization names, subject to further check of requirement to have multiple Wikipedia entities, was combined from several preliminary lists following below, with entities as they were at May 2021. 

Top 40 companies by revenues
\footnote{\url{List_of_largest_companies_by_revenue}}: 
'Walmart', 'Sinopec Group', 'Amazon', 'State Grid', 'China National Petroleum', 'Royal Dutch Shell', 'Saudi Aramco', 'Volkswagen', 'BP', 'Toyota', 'Apple', 'ExxonMobil', 'CVS Health', 'Berkshire Hathaway', 'UnitedHealth', 'McKesson', 'Glencore', 'China State Construction', 'Samsung Electronics', 'Daimler', 'Ping An Insurance', 'Alphabet', 'AT\&T', 'AmerisourceBergen', 'ICBC', 'Total', 'Foxconn', 'Trafigura', 'Exor', 'China Construction Bank', 'Ford', 'Cigna', 'Costco', 'AXA', 'Agricultural Bank of China', 'Chevron', 'Cardinal Health', 'Microsoft', 'JPMorgan Chase', 'Honda'.

Top 14 largest employers \footnote{\url{List_of_largest_employers}}: 
'U.S. Department of Defense', "People's Liberation Army", 'Walmart', 'Russian Armed Forces', "McDonald's", 'National Health Service', 'China National Petroleum Corporation', 'State Grid Corporation of China', 'Indian Railways', 'Indian Armed Forces', "Korean People's Army", 'Foxconn', 'French Ministry of National Education', 'Amazon'.

Top 10 wealthiest charitable foundations\footnote{\url{List_of_wealthiest_charitable_foundations}}:
'Novo Nordisk Foundation', 'Bill \& Melinda Gates Foundation', 'Stichting INGKA Foundation', 'Wellcome Trust', 'Howard Hughes Medical Institute', 'Azim Premji Foundation', 'Garfield Weston Foundation', 'Lilly Endowment', 'Ford Foundation', 'Silicon Valley Community Foundation'.

Top 10 largest political parties and their acronyms\footnote{\url{List_of_largest_political_parties}}:
'Bharatiya Janata Party', 'Communist Party of China', 'Democratic Party', 'Republican Party', 'Justice and Development Party', 'Aam Aadmi Party', 'Pakistan Tehreek-e-Insaf', 'Chama Cha Mapinduzi', 'United Socialist Party of Venezuela', "Cambodian People's Party".

And some of their acronyms:
'BJP', 'CPC', 'CCP', 'AKP', 'AAP', 'TPI', 'CCM', 'CPP'.

A few well known universities:
'Harvard University', 'Stanford University', 'University of Cambridge', 'Massachusetts Institute of Technology', 'University of California, Berkeley', 'Princeton University', 'Columbia University', 'California Institute of Technology', 'University of Oxford', 'University of Chicago'

And some of their short names and acronyms:
'Harvard', 'Stanford', 'Cambridge', 'Berkeley', 'Princeton', 'Columbia', 'Oxford', 'Chicago', 'Cal', 'U of C'.

Obviously many of these names did not pass the ambiguity requirement of having multiple wikipedia entities.

\section{Good sentences}\label{sec:Good_sentences}
In Section \ref{sec:Selection_for_labeling_entities} we used a notion of a 'good sentence' for sake of the Entities dataset filtering. In this context, a 'good sentence' is a sentence satisfying the following requirements:
\begin{enumerate}[topsep=0pt,itemsep=-1ex,partopsep=1ex,parsep=1ex]
    \item The sentence does not contain the strings '==' and newlines.
    \item The sentence has the number of words not less than 4, and the number of characters between 20 and 1000.
    \item The first word of the sentence is 'alpha', i.e. it consists of the alphabetical characters.
    \item The fraction of the 'alpha' words in the sentence is not lower than 0.6.
\end{enumerate}

\section{Filtering of news}\label{sec:News_filter}
The news texts obtained by the initial search (as explained in Section \ref{sec:news}) went through the initial filtering as follows:
\begin{enumerate}[topsep=0pt,itemsep=-1ex,partopsep=1ex,parsep=1ex]
    \item The name of the mention in the text must strongly overlap with at least some entity from the Entities dataset.
    \item The number of characters [‘*’, ‘\#’, ‘\&’] in the text must be not higher than 2. The reason is that these characters are often used for bullets, and the fraction of varied listings (sport, TV, business) in random news is high.
    \item Fraction of non-alpha characters in the text must be not higher than 0.3. The reason, similar to above, is to remove business documents or texts filled with sport scores.
\end{enumerate}
This still left more than half million news texts (each with a mention of interest). Additional filtering is done for sake of manageable labeling: (1) keeping only texts with keeping only texts the length of between 500 and 3000 characters, and (2) keeping only texts with mentions that can be confused with only 3-10 Wikipedia entities. This filtering step still left more than 17K news texts. Finally, the number was reduced to the 'best quality' 1000 news texts by the following filtering:
\begin{enumerate}[topsep=0pt,itemsep=-1ex,partopsep=1ex,parsep=1ex]
    \item Removing texts with containing the strings: 'wedding', 'service will be held', 'leaves to cherish', 'died peacefully', 'marriage', 'annual', 'passed away', 'survived by', 'preceded in death'. The reason for this is to reduce the fraction of names that do not exist in the Entities dataset but have the same or similar names. Having such samples is important, but we want also to have a good fraction of entities from random news that do belong to our Entities dataset.
    \item Removing texts which had, after NLTK sentence tokenization, average length of sentence less than 60 characters. The reason is to remove texts containing long listings; a normal average news sentence is not so short.
    \item Removing texts with fraction of newline characters higher than 1\%.
    \item Removing texts containing the strings: 'http', 'www.', '.com', '**', '[', '\{', '\@' or '\#'. The reason is to filter out texts with advertisements, and texts on obscure subjects from obscure sources.  
    \item Sorting the remaining (less than 2000) texts by count of non-alpha and non-digit symbols and selecting the 1000 cleanest. 
\end{enumerate}

\section{Filtering of backlinks}\label{sec:Backlinks_filter}
The filtering and cleaning of backlinks texts (Section \ref{sec:Backlinks}) is done as the following. 
\begin{enumerate}[topsep=0pt,itemsep=-1ex,partopsep=1ex,parsep=1ex]
    \item Removing Wikipedia texts that have titles starting with: 'Book', 'Category', 'Draft', 'File', 'Help', 'List', 'MediaWiki', 'Portal', 'Special', 'Talk', 'Template', 'User', 'WikiProject', 'Wikipedia'.   
    \item Cutting any Wikipedia text starting from any section named as: 'Bibliography', 'Discography', 'External Links', 'Filmography', 'Footnotes', 'Further Reading', 'Notes', 'References', 'See Also'.
    \item Removing all mentions occurring in 'bad' text locations. A 'bad' location means that a piece of text in between 100 characters before and 100 characters after the mention does not satisfy the following conditions:
    \begin{enumerate}[topsep=0pt,itemsep=-1ex,partopsep=1ex,parsep=1ex]
        \item Fraction of digits must not exceed max 0.05 of all the characters.
        \item Fraction of alpha-characters must not be below min 0.8.
        \item Fraction of newlines must not exceed max 0.01.
    \end{enumerate}
    \item Cutting the text starting from 1000 characters down from the lowest occurring mention.
\end{enumerate}
The 'removing' of a mention means that the mention looses its link notations, and becomes simple word (or words) in the text.

\section{Mention-Entities Ambiguity}\label{sec:Overlaps}
\subsection{Entities}\label{sec:Overlaps_Entities}
In order to characterize ambiguity of the named entities in a dataset, we suggested and described a 'mention-entities overlap' in Section \ref{sec:DatasetFeatures}. For clarity, here we define the overlap. 

For a single dataset, for example Entities, as in Section \ref{sec:DatasetFeatures_Entities}, we have $M$ entities, each entity is having its own text $T_i, i=1,2,...,M$.
Each text $T_i$ has a list $T_i^{(s)}$ of mentions $u_{\alpha}$ that were tagged "Same", and a list $T_i^{(o)}$ of mentions $v_{\alpha}$ that were tagged "Other". 
In each "Same" list $T_i^{(s)}$ we will also include the 'official' name of the entity (the title of the corresponding Wikipedia page) $t_i$.
For example, the "Same" lists of the first two texts: 
\begin{equation} \label{eq:Texts_list_same_1}
T_1^{(s)} = [t_1, u_1, ..., u_{N_1^{(s)}}]
\end{equation}
\begin{equation} \label{eq:Texts_list_same_2}
T_2^{(s)} = [t_2, u_{N_1^{(s)}+1}, ..., u_{N_2^{(s)}}]
\end{equation}
Similarly, the "Other" lists of mentions:
\begin{equation} \label{eq:Texts_list_other}
T_i^{(o)} = [v_{N_{i-1}^{(o)}+1}, ..., v_{N_i}^{(o)}]
\end{equation}
Here $N_0^{(o)}=0$ and $N_0^{(s)}=0$. Of course some "Other" lists can be empty.

For example, our entity "Milton, Indiana" has three mentions in its text: "Milton" (Same), "Milton" (Other), "Milton" (Same). Its list $T^{(s)}$ is ["Milton, Indiana", "Milton", "Milton"], and its list $T^{(o)}$ is ["Milton"]. The entity "David Jones (footballer, born 1940)" has eight mentions, first one is "David Willmott Llewellyn Jones", and the other seven are all "Jones". Since all these mentions are tagged "Same", they all (and the title) go into $T^{(s)}$, and $T^{(o)}$ is empty.

We define the \textbf{mention-entities ambiguity} $C^{(s)}$ for the "Same" mentions as:
\begin{equation} \label{eq:Overlap_Same}
C^{(s)} = \frac{1}{N_M^{(s)}} \sum_{\alpha=1}^{N_M^{(s)}} |\{i: u_{\alpha} \in T_i^{(s)}\}|
\end{equation}
Here $|\{i: u_{\alpha} \in T_i^{(s)}\}|$ is the number of all the entities $i$ that had at least one mention $u_{\alpha}$ (aka string, 'surface form' occurrence) tagged "Same".
Similarly, the \textbf{mention-entities ambiguity} $C^{(o)}$ for the "Other" mentions is:
\begin{equation} \label{eq:Overlap_Other}
C^{(o)} = \frac{1}{N_M^{(o)}} \sum_{\alpha=1}^{N_M^{(o)}} |\{i: v_{\alpha} \in T_i^{(s)}\}|
\end{equation}
Generally, for any dataset that has known correct mentions of entities (in our Entities dataset they are tagged as "Same"), and for any kind $o$ of mentions $v_{\alpha}$ of interest, we can define the ambiguity by the equation \ref{eq:Overlap_Other}.

In our Entities dataset, the mention-entities ambiguity is 181.3 for "Same", and 20.9 for "Other".

\subsection{News and Backlinks to Entities}\label{sec:Overlaps_News}
In Appendix \ref{sec:Overlaps_Entities} we considered ambiguity of mentions from the same dataset (e.g. Entities). Here we are considering ambiguity of mentions in one dataset $D_1$ (in our case News or Backlinks) with respect to entities of another dataset $D_2$ (in our case Entities).

We can apply the same definition by the equation \ref{eq:Overlap_Other}, with an understanding that the lists $T_i^{(s)}$ are from $D_2$, but the summation is done over the mentions $v_{\alpha}$ from $D_1$:
\begin{equation} \label{eq:Overlap_News}
C^{(o)} = \frac{1}{N^{(o)}} \sum_{\alpha=1}^{N^{(o)}} |\{i: v_{\alpha} \in T_i^{(s)}\}|
\end{equation}
Here $N^{(o)}$ is simply the number of the mentions of interest in $D_1$. 

For the 724 News mentions that were not tagged as one of entities of Entities, the ambiguity $C^{(o)} = 1.6$. A caveat here is that, while the averaging in the equation \ref{eq:Overlap_News} is done over all the 724 mentions, the ambiguity is actually created by 77\% of these mentions (555 mentions) that do exist in one or more lists $T_i^{(s)}$. The remaining 23\%, while having similar names, do not exactly match the mentions of the Entities texts. 

For the 276 News mentions that were tagged as the entities of Entities, we have to add each mention to the list $T_i^{(s)}$ of the entity with which the mention was linked (by the annotators). With or without taking this into account, the ambiguity $C^{(o)} = 1.9$ (the difference is only about 2\%). 

For the 29019 Backlinks mentions the ambiguity $C^{(o)} = 7.6$.

\end{document}